\documentclass{article}

\usepackage{microtype}
\usepackage{graphicx}
\usepackage{subfigure}
\usepackage{booktabs} % for professional tables
\usepackage{multirow}
\usepackage[normalem]{ulem}

\usepackage{listings}
\usepackage{courier}

\usepackage{hyperref}

\newcommand{\ms}[2]{#1\tiny{$\pm$#2}}

\usepackage{amsmath,amsfonts,bm}

\def\1{\bm{1}}

\def\vtheta{{\bm{\theta}}}

\def\vu{{\bm{u}}}
\def\vv{{\bm{v}}}
\def\vw{{\bm{w}}}
\def\vx{{\bm{x}}}

\def\vz{{\bm{z}}}

\DeclareMathAlphabet{\mathsfit}{\encodingdefault}{\sfdefault}{m}{sl}
\SetMathAlphabet{\mathsfit}{bold}{\encodingdefault}{\sfdefault}{bx}{n}

\newcommand{\KL}{D_{\mathrm{KL}}}

\usepackage[accepted]{icml2020}

\icmltitlerunning{
Self-supervised Label Augmentation
via Input Transformations
}

\begin{document}

\twocolumn[
\icmltitle{
Self-supervised Label Augmentation
via Input Transformations
}
%Label Augmentation} % with Input Transformations}

% It is OKAY to include author information, even for blind
% submissions: the style file will automatically remove it for you
% unless you've provided the [accepted] option to the icml2020
% package.

% List of affiliations: The first argument should be a (short)
% identifier you will use later to specify author affiliations
% Academic affiliations should list Department, University, City, Region, Country
% Industry affiliations should list Company, City, Region, Country

% You can specify symbols, otherwise they are numbered in order.
% Ideally, you should not use this facility. Affiliations will be numbered
% in order of appearance and this is the preferred way.
\icmlsetsymbol{equal}{*}

\begin{icmlauthorlist}
\icmlauthor{Hankook Lee}{kaist_ee}
\icmlauthor{Sung Ju Hwang}{kaist_ai,kaist_cs,aitrics}
\icmlauthor{Jinwoo Shin}{kaist_ai,kaist_ee}
\end{icmlauthorlist}

\icmlaffiliation{kaist_ee}{School of Electrical Engineering, KAIST, Daejeon, Korea}
\icmlaffiliation{kaist_ai}{Graduate School of AI, KAIST, Daejeon, Korea}
\icmlaffiliation{kaist_cs}{School of Computing, KAIST, Daejeon, Korea}
\icmlaffiliation{aitrics}{AITRICS, Seoul, Korea}

\icmlcorrespondingauthor{Jinwoo Shin}{jinwoos@kaist.ac.kr}

% You may provide any keywords that you
% find helpful for describing your paper; these are used to populate
% the "keywords" metadata in the PDF but will not be shown in the document
\icmlkeywords{Machine Learning, ICML}

\vskip 0.3in
]

% this must go after the closing bracket ] following \twocolumn[ ...

% This command actually creates the footnote in the first column
% listing the affiliations and the copyright notice.
% The command takes one argument, which is text to display at the start of the footnote.
% The \icmlEqualContribution command is standard text for equal contribution.
% Remove it (just {}) if you do not need this facility.

\printAffiliationsAndNotice{}  % leave blank if no need to mention equal contribution
% \printAffiliationsAndNotice{\icmlEqualContribution} % otherwise use the standard text.

\begin{abstract}
Self-supervised learning, which learns by constructing artificial labels given only the input signals, has recently gained considerable attention for learning representations with unlabeled datasets, i.e., learning without any human-annotated supervision. In this paper, we show that such a technique can be used to significantly improve the model accuracy even under fully-labeled datasets. Our scheme trains the model to learn both original and self-supervised tasks, but is different from conventional multi-task learning frameworks that optimize the summation of their corresponding losses. Our main idea is to learn a single unified task with respect to the joint distribution of the original and self-supervised labels, i.e., we augment original labels via self-supervision of input transformation. This simple, yet effective approach allows to train models easier by relaxing a certain invariant constraint during learning the original and self-supervised tasks simultaneously. It also enables an aggregated inference which combines the predictions from different augmentations to improve the prediction accuracy. Furthermore, we propose a novel knowledge transfer technique, which we refer to as self-distillation, that has the effect of the aggregated inference in a single (faster) inference. We demonstrate the large accuracy improvement and wide applicability of our framework on various fully-supervised settings, e.g., the few-shot and imbalanced classification scenarios.
\end{abstract}

\section{Introduction}

In recent years, \emph{self-supervised learning}~\cite{doersch2015context} has shown remarkable success in unsupervised representation learning for images~\cite{doersch2015context,noroozi2016jigsaw,larsson2017colorization,gidaris2018rotation,zhang2019aet}, natural language~\cite{devlin2018bert}, and video games~\cite{anand2019unsupervised_atari}. When human-annotated labels are scarce, the approach constructs artificial labels, referred to as \emph{self-supervision}, only using the input examples and then learns their representations via predicting the labels. One of the simplest, yet effective self-supervised learning approaches is to predict which transformation $t$ is applied to an input $\vx$ from observing only the modified input $t(\vx)$, e.g., $t$ can be a patch permutation \cite{noroozi2016jigsaw} or a rotation \cite{gidaris2018rotation}. To predict such transformations, a model should distinguish between what is semantically natural or not, and consequently, it learns high-level semantic representations of inputs.

The simplicity of transformation-based self-supervision has encouraged its wide applicability for other purposes beyond unsupervised representation learning, e.g., semi-supervised learning~\citep{zhai2019s4l,berthelot2020remixmatch}, improving robustness~\cite{hendrycks2019self_robustness}, and training generative adversarial networks~\cite{chen2018self_gan}. The prior works commonly maintain two separate classifiers (yet sharing common feature representations) for the original and self-supervised tasks, and optimize their objectives simultaneously. However, this multi-task learning approach typically provides no accuracy gain when working with fully-labeled datasets. This inspires us to explore the following question: \emph{how can we effectively utilize the transformation-based self-supervision for fully-supervised classification tasks?}

\begin{figure*}[t]
\centering
\hspace*{\fill}
\subfigure[Difference with previous approaches\label{fig:overview:diff}]{\includegraphics[scale=0.53]{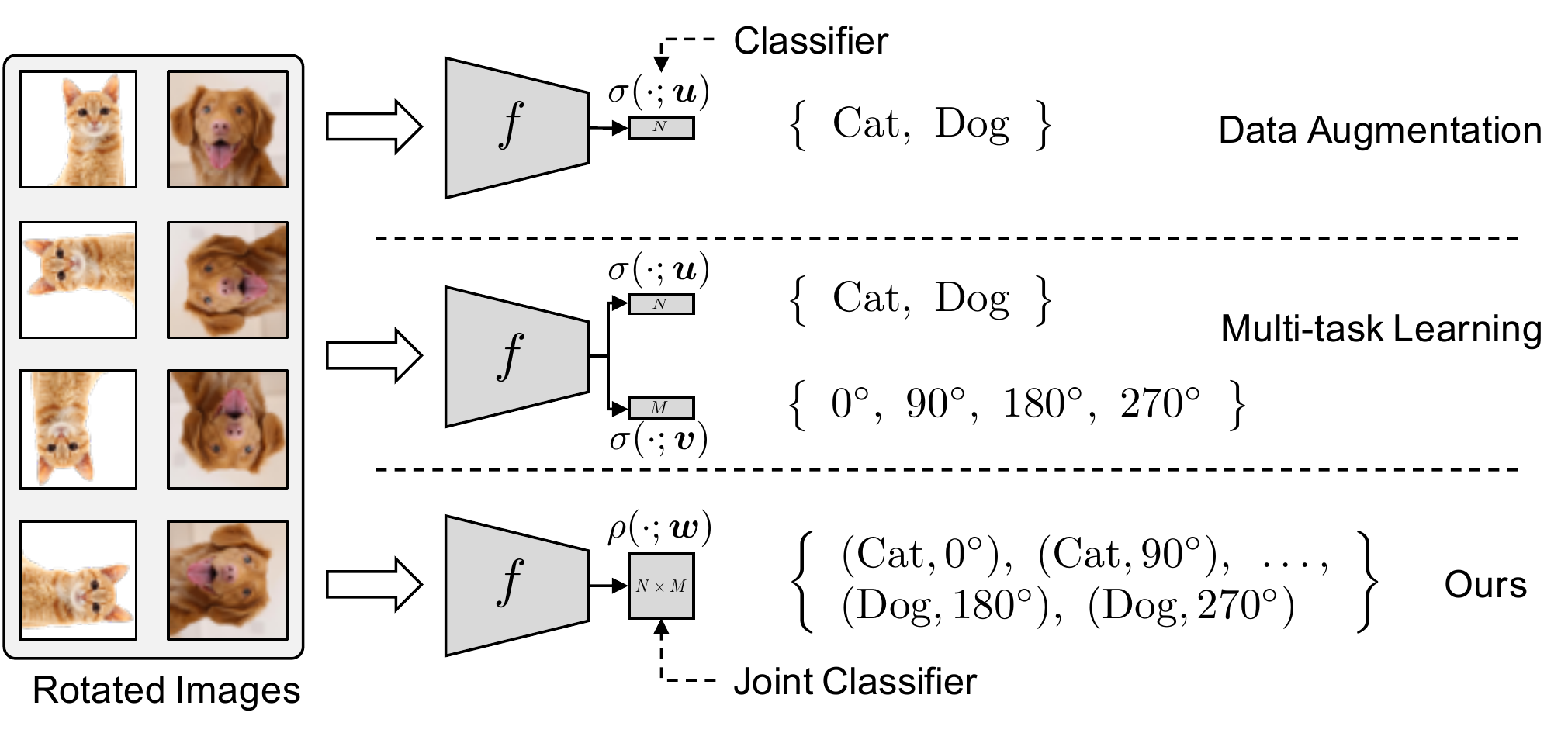}}
\hspace*{\fill}
\subfigure[Aggregation \& self-distillation\label{fig:overview:agg}]{\includegraphics[scale=0.53]{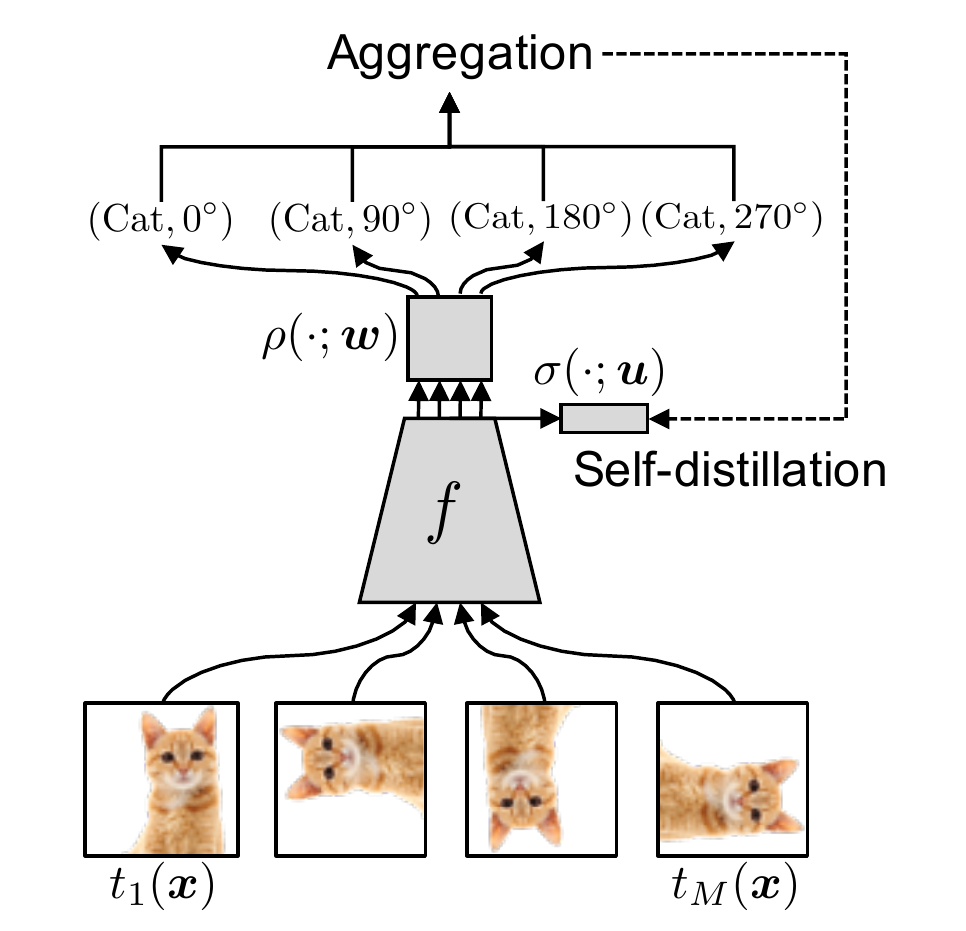}}
\hspace*{\fill}
\\
\hspace*{\fill}
\subfigure[Rotation ($M=4$)\label{fig:overview:rot_ex}]{\includegraphics[scale=0.44]{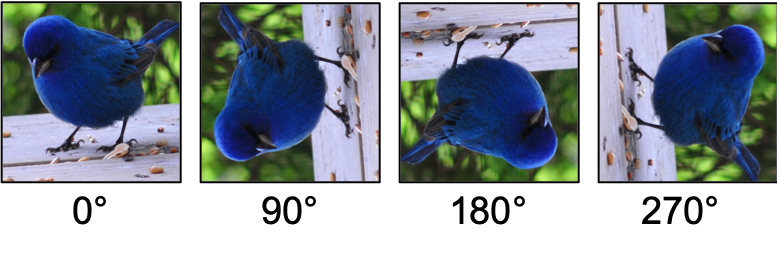}}
\hspace*{\fill}
\subfigure[Color permutation ($M=6$)\label{fig:overview:cp_ex}]{\includegraphics[scale=0.44]{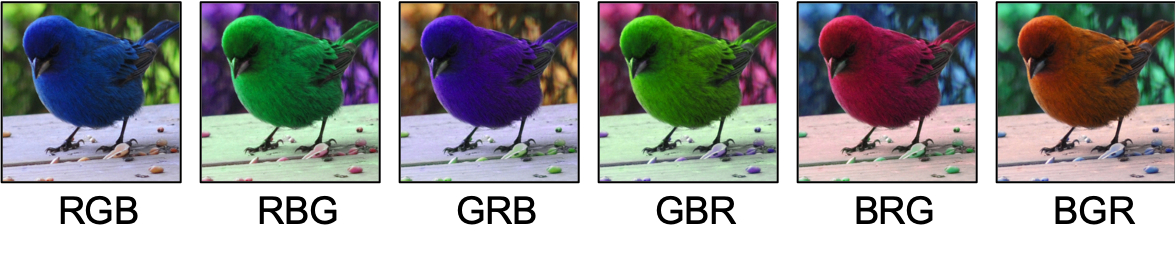}}
\hspace*{\fill}
\caption{(a) An overview of our self-supervised label augmentation and previous approaches with self-supervision.
(b) Illustrations of our aggregation method utilizing all augmented samples and self-distillation method transferring the aggregated knowledge
into itself.
(c) Rotation-based augmentation. (d) Color-permutation-based augmentation.
}\label{fig:overview}
\vspace{-0.1in}
\end{figure*}

\textbf{Contribution.}
We first discuss our observation that the multi-task learning approach forces the primary classifier for the original task to be invariant with respect to transformations of a self-supervised task. For example, when using rotations as self-supervision~\cite{zhai2019s4l}, which rotates each image $0$, $90$, $180$, $270$ degrees while preserving its original label, the primary classifier is forced to learn representations that are invariant to the rotations. Forcing such invariance could lead to increasing complexity of tasks since the transformations could largely change characteristics of samples and/or meaningful information for recognizing objects, e.g., image classification \{6 vs.\ 9\} or \{bird vs.\ bat\}.\footnote{This is because bats hang typically upside down, while birds do not.} Consequently, this could hurt the overall representation learning, and degrade the classification accuracy of the primary fully-supervised model (see Table \ref{table:ablation} in Section \ref{sec:exp:ablation}).

To tackle this challenge, we propose a simple yet effective idea (see Figure \ref{fig:overview:diff}), which is to learn a single unified task with respect to the \emph{joint} distribution of the original and self-supervised labels, instead of two separate tasks typically used in the prior self-supervision literature. For example, when training on CIFAR10~\cite{krizhevsky2009cifar} (10 labels) with the self-supervision on rotation (4 labels), we learn the {joint} probability distribution on all possible combinations, i.e., 40 labels.

This label augmentation method, which we refer to as \emph{self-supervised label augmentation} (SLA), does not force any invariance to the transformations without assumption for the relationship between the original and self-supervised labels. Furthermore, since we assign {different self-supervised labels} for each transformation, it is possible to make a prediction by {\emph{aggregation}} across all transformations at test time, as illustrated in Figure \ref{fig:overview:agg}. This can provide an (implicit) ensemble effect using a single model. Finally, to speed up the inference process without loss of the ensemble effect, we propose a novel \emph{self-distillation} technique that transfers the knowledge of the multiple inferences into a single inference, as illustrated in Figure \ref{fig:overview:agg}.

In our experiments, we consider two types of input transformations for self-supervised label augmentation, \emph{rotation} (4 transformations) and \emph{color permutation} (6 transformations), as illustrated in Figure \ref{fig:overview:rot_ex} and Figure \ref{fig:overview:cp_ex}, respectively. To demonstrate the wide applicability and compatibility of our method, we experiment with various benchmark datasets and classification scenarios, including the few-shot and imbalanced classification tasks. In all tested settings, our simple method improves the classification accuracy significantly and consistently. For example, our method achieves $8.60\%$ and $7.05\%$ relative accuracy gains on the standard fully-supervised task on CIFAR-100 \cite{krizhevsky2009cifar} and the 5-way 5-shot task on FC100~\cite{oreshkin2018tadam}, respectively, over relevant baselines.\footnote{Code available at \url{https://github.com/hankook/SLA}.}

\section{Self-supervised Label Augmentation}\label{sec:method}

In this section, we provide the details of our self-supervised label augmentation techniques under focusing on the fully-supervised scenarios. We first discuss the conventional multi-task learning approach utilizing self-supervised labels and its limitations in Section \ref{sec:method:pre}. Then, we introduce our learning framework which can fully utilize the power of self-supervision in Section \ref{sec:method:joint}. Here, we also propose two additional techniques: \emph{aggregation}, which utilizes all {differently} augmented samples for providing an ensemble effect using a single model; and \emph{self-distillation}, which transfers the aggregated knowledge into the model itself for accelerating the inference speed without loss of the ensemble effect.

\textbf{Notation.} Let $\vx \in \mathbb{R}^d$ be an input, $y\in\{1,\ldots,N\}$ be its label where $N$ is the number of classes, $\mathcal{L}_\text{CE}$ be the cross-entropy loss function, $\sigma(\cdot;\vu)$ be the softmax classifier, i.e., $\sigma_i(\vz;\vu)=\exp(\vu_i^\top\vz)/\sum_k\exp(\vu_k^\top\vz)$, and $\vz=f(\vx;\vtheta)$ be an embedding vector of $\vx$ where $f$ is a neural network with the parameter $\vtheta$. We also let $\tilde{\vx}=t(\vx)$ denote an augmented sample using a transformation $t$, and $\tilde{\vz}=f(\tilde{\vx};\vtheta)$ be the
embedding of the augmented sample $\tilde{\vx}$.

\subsection{{Multi-task Learning with Self-supervision}}\label{sec:method:pre}

In {transformation-based} self-supervised learning~\cite{doersch2015context,noroozi2016jigsaw,larsson2017colorization,gidaris2018rotation,zhang2019aet}, models learn to predict which transformation $t$ is applied to an input $\vx$ given a modified sample $\tilde{\vx}=t(\vx)$. The common approach to utilize self-supervised labels for other task is to optimize two losses of the primary and self-supervised tasks, while sharing the feature space among them \cite{chen2018self_gan,hendrycks2019self_robustness,zhai2019s4l}; that is, the two tasks are trained in a multi-task learning framework. Thus, in the fully-supervised setting, one can formulate the multi-task objective $\mathcal{L}_\text{MT}$ with self-supervision as follows:
\begin{align}
    & \mathcal{L}_\text{MT}(\vx,y;\vtheta,\vu,\vv)  \nonumber \\
    & = \frac{1}{M}\sum_{j=1}^M\mathcal{L}_\text{CE}(\sigma(\tilde{\vz}_j;\vu),y)
     +\mathcal{L}_\text{CE}(\sigma(\tilde{\vz}_j;\vv),j), \label{eq:loss_mt}
\end{align}
where $\{t_j\}_{j=1}^M$ is pre-defined transformations, $\tilde \vx_j=t_j(\vx)$ is a transformed sample by $t_j$, and $\tilde \vz_j=f(\tilde \vx_j;\vtheta)$ is its embedding of the neural network $f$. Here, $\sigma(\cdot;\vu)$ and $\sigma(\cdot;\vv)$ are classifiers for primary and self-supervised tasks, respectively. The above loss forces the primary classifier $\sigma(f(\cdot);\vu)$ to be invariant to the transformations $\{t_j\}$. Depending on the type of transformations, forcing such invariance may not make sense, as the statistical characteristics of the augmented training samples (e.g., via rotation) could become very different from those of {original} training samples. In such a case, enforcing invariance to those transformations would make the learning more difficult, and can even degrade the performance (see Table \ref{table:ablation} in Section \ref{sec:exp:ablation}).

In the multi-task learning objective \eqref{eq:loss_mt}, if we do not learn self-supervision, then it can be considered as a data augmentation objective $\mathcal{L}_\text{DA}$ as follows:
\begin{align}
    \mathcal{L}_\text{DA}(\vx,y;\vtheta,\vu)
    & = 
    \frac{1}{M}\sum_{j=1}^M \mathcal{L}_\text{CE}(\sigma(\tilde{\vz}_j;\vu),y).
    \label{eq:loss_da}
\end{align}
This conventional data augmentation aims to improve upon the generalization ability of the target neural network $f$ by leveraging certain transformations that can preserve their semantics, e.g., cropping, contrast enhancement, and flipping. On the other hands, if a transformation modifies the semantics, the invariant property with respect to the transformation could interfere with semantic representation learning (see Table \ref{table:ablation} in Section \ref{sec:exp:ablation}).

\subsection{{Eliminating Invariance via Joint-label Classifier}}\label{sec:method:joint}

Our key idea is to remove the {unnecessary} invariant property of the classifier $\sigma(f(\cdot);\vu)$ in \eqref{eq:loss_mt} and \eqref{eq:loss_da} among the transformed samples. To this end, we use a joint softmax classifier $\rho(\cdot;\vw)$ which represents the joint probability as $P(i,j|\tilde{\vx})=\rho_{ij}(\tilde{\vz};\vw)=\exp(\vw_{ij}^\top\tilde{\vz})/\sum_{k,l}\exp(\vw_{kl}^\top\tilde{\vz})$. Then, our training objective can be written as
\begin{align}\label{eq:loss_sda}
    \mathcal{L}_\text{SLA}(\vx,y;\vtheta,\vw)=\frac{1}{M}\sum_{j=1}^M\mathcal{L}_\text{CE}(\rho(\tilde{\vz}_j;\vw),(y,j)),
\end{align}
where $\mathcal{L}_\text{CE}(\rho(\tilde{\vz};\vw),(i,j))=-\log \rho_{ij}(\tilde{\vz};\vw)$. Note that this framework only increases the number of labels, thus the number of additional parameters is negligible compared to that of the whole network, e.g., only 0.4\% parameters are newly introduced when using ResNet-32 \cite{he2016resnet}. We also remark that the above objective can be reduced to the multi-task learning objective $\mathcal{L}_\text{MT}$ \eqref{eq:loss_mt} when $\vw_{ij}=\vu_i+\vv_j$ for all $i,j$, and the data augmentation objective $\mathcal{L}_\text{DA}$ \eqref{eq:loss_da} when $\vw_{ij}=\vu_i$ for all $i$. From the perspective of optimization, $\mathcal{L}_\text{MT}$ and $\mathcal{L}_\text{SLA}$ consider the same set of multi-labels, but the former requires the additional constraint, thus it is harder to optimize than the latter. The difference between the conventional augmentation, multi-task learning and ours is illustrated in Figure \ref{fig:overview:diff}. During training, we feed all $M$ augmented samples simultaneously for each iteration as \citet{gidaris2018rotation} did, i.e., we minimize $\frac{1}{|B|}\sum_{(\vx,y)\in B}\mathcal{L}_\text{SLA}(\vx,y;\vtheta,\vw)$ for each mini-batch $B$. We also assume that the first transformation is the identity function, i.e., $\tilde{\vx}_1=t_1(\vx)=\vx$.

\textbf{Aggregated inference.} Given a test sample $\vx$ or its augmented sample $\tilde{\vx}_j=t_j(\vx)$ by a transformation $t_j$, we do not need to consider all $N\times M$ labels for the prediction of its original label, because we already know which transformation is applied. Therefore, we make a prediction using the conditional probability $P(i|\tilde{\vx}_j,j)=\exp(\vw_{ij}^\top\tilde{\vz}_j)/\sum_k\exp(\vw_{kj}^\top\tilde{\vz}_j)$ where $\tilde{\vz}_j=f(\tilde{\vx}_j)$. Furthermore, for all possible transformations $\{t_j\}$, we aggregate the corresponding conditional probabilities to improve the classification accuracy, i.e., we train a single model, which can perform inference like an ensemble model. To compute the probability of the \emph{aggregated inference}, we first average pre-softmax activations {(i.e., logits)}, and then compute the softmax probability as follows:
\begin{align}
P_\text{aggregated}(i|\vx) & = \frac{\exp(s_i)}{\sum_{k=1}^N\exp(s_k)}, \label{eq:agg}
\end{align}
where $s_i=\frac{1}{M}\sum_{j=1}^M\vw_{ij}^\top\tilde{\vz}_{j}$. Since we assign different labels for each transformation $t_j$, our aggregation scheme improves accuracy significantly. Somewhat surprisingly, it achieves comparable performance with the ensemble of multiple independent models in our experiments (see Table \ref{table:compare} in Section \ref{sec:exp:ablation}). We refer to the counterpart of the aggregation as \emph{single inference}, which uses only the non-augmented or original sample $\tilde{\vx}_1=\vx$, i.e., predicts a label using $P(i|\tilde{\vx}_1,j{=}1)=\exp(\vw_{i1}^\top f(\vx;\vtheta))/\sum_k\exp(\vw_{k1}^\top f(\vx;\vtheta))$.

\textbf{Self-distillation from aggregation.} Although the aforementioned aggregated inference achieves outstanding performance, it requires to compute $\tilde{\vz}_j=f(\tilde{\vx}_j)$ for all $j$, i.e., it requires $M$ times higher computation cost than the single inference. To accelerate the inference, we perform self-distillation \cite{hinton2015knowledge_distillation,lan2018one} from the aggregated knowledge $P_\text{aggregated}(\cdot|\vx)$ to another classifier $\sigma(f(\vx;\vtheta);\vu)$ parameterized by $\vu$, as illustrated in Figure \ref{fig:overview:agg}. Then, the classifier $\sigma(f(\vx;\vtheta);\vu)$ can maintain the aggregated knowledge using only one embedding $\vz=f(\vx)$. To this end, we optimize the following objective:
\begin{align}
    \mathcal{L}_\text{SLA+SD}(\vx,y;\vtheta,\vw,\vu) & =\mathcal{L}_\text{SLA}(\vx,y;\vtheta,\vw)\nonumber\\
    & \quad + \KL(P_\text{aggregated}(\cdot|\vx) \Vert \sigma(\vz;\vu)) \nonumber \\
    & \quad + \beta\mathcal{L}_\text{CE}(\sigma(\vz;\vu),y),
\label{eq:loss_sda_sd}
\end{align}
where $\beta$ is a hyperparameter and we simply choose $\beta\in\{0,1\}$. When computing the gradient of $\mathcal{L}_\text{SLA+SD}$, we consider $P_\text{aggregated}(\cdot|\vx)$ as a constant. After training, we use $\sigma(f(\vx;\vtheta);\vu)$ for inference without aggregation.

\section{Experiments}\label{sec:exp}

\begin{figure*}[t]
\centering
\subfigure[Upright images\label{fig:upright}]{\includegraphics[scale=0.27]{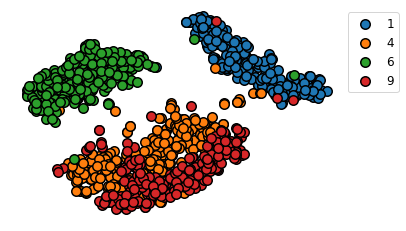}}
\hspace*{\fill}
\subfigure[1 vs.~9 rotated images\label{fig:1vs9_rotated}]{\includegraphics[scale=0.27]{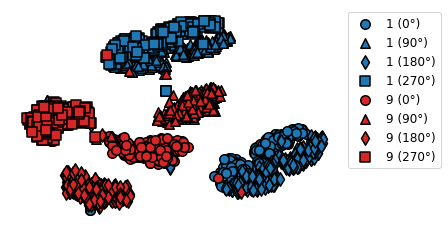}}
\subfigure[4 vs.~9 rotated images\label{fig:4vs9_rotated}]{\includegraphics[scale=0.27]{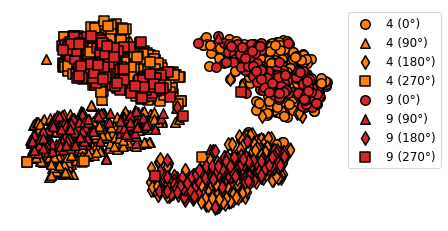}}
\subfigure[6 vs.~9 rotated images\label{fig:6vs9_rotated}]{\includegraphics[scale=0.27]{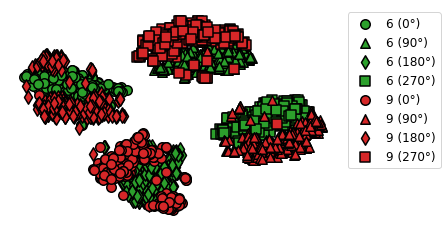}}
\caption{Visualization of raw pixels of 1, 4, 6 and 9 in MNIST \cite{lecun1998mnist} by t-SNE
\cite{maaten2008tsne}. Colors and shapes indicate digits and rotation, respectively.}\label{fig:example}
\vspace{-0.2in}
\end{figure*}

We experimentally validate our self-supervised label augmentation techniques described in Section \ref{sec:method}.
Throughout this section, we refer to 
data augmentation $\mathcal{L}_\text{DA}$ \eqref{eq:loss_da} as DA, 
multi-task learning $\mathcal{L}_\text{MT}$ \eqref{eq:loss_mt} as MT,
and our self-supervised label augmentation $\mathcal{L}_\text{SLA}$ \eqref{eq:loss_sda} as SLA for notational simplicity. %simplification.
We also refer baselines which use only random cropping and flipping for data augmentation (without rotation and color permutation) as ``Baseline''. Note that DA is different from Baseline because DA uses self-supervision as augmentation (e.g., rotation) while Baseline does not.
After training with $\mathcal{L}_\text{SLA}$, we consider two inference schemes:
the single inference $P(i|\vx,j=1)$ and the aggregated inference $P_\text{aggregated}(i|\vx)$ denoted by SLA+SI and SLA+AG, respectively.
We also denote the self-distillation method $\mathcal{L}_\text{SLA+SD}$ \eqref{eq:loss_sda_sd} as SLA+SD which
uses only the single inference $\sigma(f(\vx;\vtheta);\vu)$.

\subsection{Setup}\label{sec:exp:setup}

\textbf{Datasets and models.}
We evaluate our method on various classification datasets:
CIFAR10/100 \cite{krizhevsky2009cifar}, Caltech-UCSD Birds or CUB200 \cite{WahCUB_200_2011},
Indoor Scene Recognition or MIT67 \cite{quattoni2009indoor},
Stanford Dogs \cite{Khosla2011dogs}, and tiny-ImageNet\footnote{
\url{https://tiny-imagenet.herokuapp.com/}}
for standard or imbalanced image classification;
mini-ImageNet \cite{vinyals2016matching},
CIFAR-FS \cite{bertinetto2018r2d2}, and FC100 \cite{oreshkin2018tadam}
for few-shot classification.
Note that CUB200, MIT67, and Stanford Dogs are fine-grained datasets.
{We use 
32-layer residual networks \cite{he2016resnet} for CIFAR and
18-layer residual networks for three fine-grained datasets and tiny-ImageNet
unless otherwise stated.}

\textbf{Implementation details.} %See Appendix.
For the standard image classification datasets, we use SGD with a learning rate of $0.1$, momentum of $0.9$, and weight decay of $10^{-4}$.
We train for $80$k iterations with a batch size of $128$. For the fine-grained datasets, we train for $30$k iterations with a batch size of $32$ because
they have a relatively smaller number of training samples.
We decay the learning rate by the constant factor of $0.1$ at $50$\% and $75$\% iterations.
We report the average accuracy of three trials for all experiments unless otherwise noted.
When combining with other methods, we use publicly available codes and follow their
experimental setups: MetaOptNet~\cite{lee2019metaoptnet} for few-shot learning,
LDAM~\cite{cao2019ldam} for imbalanced datasets, and FastAutoAugment~\cite{lim2019fastautoaugment} and CutMix~\cite{yun2019cutmix}
for advanced augmentation experiments.
In the supplementary material,
we provide pseudo-codes of our algorithm, which can be easily implemented.

\textbf{Choices of transformation.}
Since using the entire input images during training is important 
{for image classification},
some self-supervision techniques are not suitable for our purpose.
For example, the Jigsaw puzzle approach \cite{noroozi2016jigsaw} divides an input image to $3\times3$ patches and
then computes their embedding separately. Prediction using {such} embedding performs worse than that using the entire image.
To avoid this issue, we choose two transformations that use the entire input image without cropping: %preserve the continuity:
\emph{rotation}~\cite{gidaris2018rotation} and \emph{color permutation}.
Rotation constructs $M=4$ rotated images ($0^\circ$, $90^\circ$, $180^\circ$, $270^\circ$) as illustrated in Figure \ref{fig:overview:rot_ex}.
This transformation is widely used for self-supervision
due to its simplicity \cite{chen2018self_gan,zhai2019s4l}.
Color permutation constructs $M=3!=6$ different images via swapping RGB channels as illustrated in Figure \ref{fig:overview:cp_ex}.
This transformation can be useful when color information is important such as fine-grained classification datasets.

\subsection{Ablation Study}\label{sec:exp:ablation}

\textbf{Toy example for intuition.}
To provide intuition on the difficulty of
learning an invariant property with respect to certain 
transformations,
we here introduce simple examples: three binary digit-image classification tasks,
\{1 vs.~9\}, \{4 vs.~9\}, and \{6 vs.~9\} in MNIST
\cite{lecun1998mnist} using linear classifiers based on raw pixel values.
As illustrated in Figure \ref{fig:upright}, it
is often easier to classify the upright digits using a linear classifier,
e.g., 0.2\% error when classifying only upright 6s and 9s.
Note that 4 and 9 have similar shapes, so their pixel values are closer than other pairs.
After rotating digits while preserving labels, the linear classifiers
can still distinguish between rotated 1 and 9 as illustrated in
Figure \ref{fig:1vs9_rotated}, but
cannot between rotated 4, 6 and 9, as illustrated in Figure
\ref{fig:4vs9_rotated} and \ref{fig:6vs9_rotated}, e.g., 13\% error when classifying
rotated 6s and 9s.
These examples show that linear separable data could be no longer linear separable after augmentation by some
transformations such as rotation, i.e., explain why forcing an invariant property can increase the difficulty of learning tasks.
However, if assigning a different label for each rotation (as we propose in this paper), then
the linear classifier can classify the rotated digits, e.g., 1.1\% error when classifying
rotated 6s and 9s.

\begin{table}[t]
\caption{Classification accuracy (\%) of single inference using data augmentation (DA), multi-task learning (MT), and our self-supervised
label augmentation (SLA) with rotation.
The best accuracy is indicated as bold.
}\label{table:ablation}
\vskip 0.1in
\begin{center}
\small
\begin{tabular}{ccccc}
\toprule
Dataset    & Baseline & DA     & MT    & SLA+SI   
\\ \midrule                                    
CIFAR10       & 92.39 & 90.44  & 90.79  & {\bf 92.50}  \\
CIFAR100      & 68.27 & 65.73  & 66.10  & {\bf 68.68}  \\
tiny-ImageNet & 63.11 & 60.21  & 58.04  & {\bf 63.99}  \\ % \midrule
\bottomrule
\end{tabular}
\end{center}
\vspace{-0.1in}
\end{table}

\begin{figure}[t!]
\centering
\vspace{0.05in}
\includegraphics[width=0.35\textwidth]{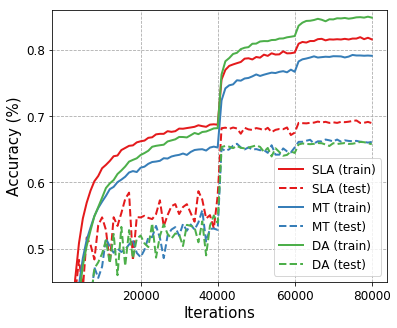}
\vspace{-0.1in}
\caption{Training curves of data augmentation (DA), multi-task learning (MT), and our self-supervised label augmentation (SLA) with rotation. The solid and dashed lines indicate training and test accuracy on CIFAR100, respectively.
}\label{fig:cifar100_training}
\vspace{-0.15in}
\end{figure}

\begin{table}[t]
\caption{
Classification accuracy (\%) of the 
independent ensemble (IE) and our aggregation using rotation (SLA+AG).
{Note that a single model
requires 0.46M parameters while four independent models do 1.86M parameters.}
The best accuracy is indicated as bold.
}\label{table:compare}
\vskip 0.1in
\begin{center}
\small
\resizebox{\columnwidth}{!}{%
\begin{tabular}{ccccc}
\toprule
              & \multicolumn{2}{c}{Single Model} & \multicolumn{2}{c}{4 Models} \\
              \cmidrule(lr){2-3}\cmidrule(lr){4-5}
Dataset       & Baseline & SLA+AG      & IE    & IE + SLA+AG \\
\midrule                       
CIFAR10       & 92.39    & {\bf 94.50} & 94.36 & {\bf 95.10} \\
CIFAR100      & 68.27    & {\bf 74.14} & 74.82 & {\bf 76.40} \\
tiny-ImageNet & 63.11    & {\bf 66.95} & 68.18 & {\bf 69.01} \\ % \midrule
\bottomrule
\end{tabular}}
\end{center}
\vspace{-0.15in}
\end{table}

\begin{table*}[t]
\caption{Classification accuracy (\%) 
{on various benchmark datasets}
using
self-supervised label augmentation
with rotation and color permutation.
SLA+SD and SLA+AG indicate the single inference trained by $\mathcal{L}_\text{SLA+SD}$, and
the aggregated inference trained by $\mathcal{L}_\text{SLA}$, respectively.
The relative gain
is shown in brackets.
}\label{table:all}
\vskip 0.1in
\begin{center}
\small
\begin{tabular}{cccccc}
\toprule
& & \multicolumn{2}{c}{Rotation} & \multicolumn{2}{c}{Color Permutation} \\
\cmidrule(lr){3-4}\cmidrule(lr){5-6}
Dataset    & Baseline & SLA+SD & SLA+AG & SLA+SD & SLA+AG   % & CUB200 & MIT67 & Stanford Dogs & ImageNet
\\ \midrule                             
CIFAR10       & 92.39 & 93.26 {\small (+0.94\%)} & 94.50 {\small (+2.28\%)}  & 91.51 {\small (-0.95\%)} & 92.51 {\small (+0.13\%)} \\
CIFAR100      & 68.27 & 71.85 {\small (+5.24\%)} & 74.14 {\small (+8.60\%)}  & 68.33 {\small (+0.09\%)} & 69.14 {\small (+1.27\%)} \\
CUB200        & 54.24 & 62.54 {\small (+15.3\%)} & 64.41 {\small (+18.8\%)}  & 60.95 {\small (+12.4\%)} & 61.10 {\small (+12.6\%)} \\
MIT67         & 54.75 & 63.54 {\small (+16.1\%)} & 64.85 {\small (+18.4\%)}  & 60.03 {\small (+9.64\%)} & 59.99 {\small (+9.57\%)} \\
Stanford Dogs & 60.62 & 66.55 {\small (+9.78\%)} & 68.70 {\small (+13.3\%)}  & 65.92 {\small (+8.74\%)} & 67.03 {\small (+10.6\%)} \\
tiny-ImageNet & 63.11 & 65.53 {\small (+3.83\%)} & 66.95 {\small (+6.08\%)}  & 63.98 {\small (+1.38\%)} & 64.15 {\small (+1.65\%)} \\
\bottomrule
\end{tabular}%}
\end{center}
\vspace{-0.2in}
\end{table*}

\begin{table}[t]
\caption{
Classification accuracy (\%) of SLA+AG based on the set (each row) of composed transformations.
We first choose subsets of rotation and color permutation (see first two columns) and compose them
where $M$ is the number of composed transformations. ALL indicates that we compose all rotations and/or
color permutations.
% We use SLA with
% the composed transformations when training models.
The best accuracy is indicated as bold.
}\label{table:comb_train}
\vskip 0.1in
\centering
\small
\begin{tabular}{cccc}
\toprule
Rotation {$T_r$}       & Color permutation {$T_c$} & $M$ & CUB200 \\
\midrule
0$^\circ$              & RGB                       & 1   & 54.24  \\
0$^\circ$, 180$^\circ$ & RGB                       & 2   & 58.92  \\
ALL                    & RGB                       & 4   & 64.41  \\
0$^\circ$              & RGB, GBR, BRG             & 3   & 56.47  \\
0$^\circ$              & ALL                       & 6   & 61.10  \\
0$^\circ$, 180$^\circ$ & RGB, GBR, BRG             & 6   & 60.87  \\
ALL                    & RGB, GBR, BRG             & 12  & \textbf{65.53}  \\
ALL                    & ALL                       & 24  & 65.43  \\
\bottomrule
\end{tabular}%}
\end{table}

\begin{table}[t]
\caption{{Classification error rates (\%) of various augmentation methods with SLA+SD on
CIFAR 10/100.
We train WideResNet-40-2 \cite{zagoruyko2016wrn} and PyramidNet200 \cite{han2017pyramidnet} following the experimental setup of 
\citet{lim2019fastautoaugment} and \citet{yun2019cutmix}, respectively.
%In the last row, we conduct only one trial. 
The best accuracy is indicated as bold.}}
\label{table:cifar-sota}
\vskip 0.1in
\small
\resizebox{\columnwidth}{!}{%
\begin{tabular}{lcc}
\toprule
& CIFAR10 & CIFAR100 \\
\midrule
WRN-40-2                               & 5.24 & 25.63 \\
+ Cutout                               & 4.33 & 23.87 \\
+ Cutout + \textbf{SLA+SD} (ours)      & 3.36 & 20.42 \\
+ FastAutoAugment                      & 3.78 & 21.63 \\
+ FastAutoAugment + \textbf{SLA+SD} (ours) & 3.06 & 19.49 \\
+ AutoAugment                          & 3.70 & 21.44 \\
+ AutoAugment + \textbf{SLA+SD} (ours) & \textbf{2.95} & \textbf{18.87} \\
\midrule
PyramidNet200                     & 3.85 & 16.45 \\
+ Mixup                           & 3.09 & 15.63 \\
+ CutMix                          & 2.88 & 14.47 \\
+ CutMix + \textbf{SLA+SD} (ours) & \textbf{1.80} & \textbf{12.24} \\
\bottomrule
\end{tabular}}
\vspace{-0.1in}
\end{table}

\textbf{Comparison with DA and MT.} We empirically verify that our proposed method can utilize self-supervision without loss of accuracy on fully-supervised datasets
while data augmentation and multi-task learning approaches cannot.
To this end, we train models on generic classification datasets, CIFAR10/100 and tiny-ImageNet, using three
different objectives: data augmentation $\mathcal{L}_\text{DA}$ (\ref{eq:loss_da}),
multi-task learning $\mathcal{L}_\text{MT}$ (\ref{eq:loss_mt}),
and our self-supervised label augmentation $\mathcal{L}_\text{SLA}$ (\ref{eq:loss_sda})
with rotation.
As reported in Table \ref{table:ablation}, $\mathcal{L}_\text{DA}$ and $\mathcal{L}_\text{MT}$
degrade the performance significantly compared to the baseline that does not use the rotation-based augmentation.
However, when training with $\mathcal{L}_\text{SLA}$, the performance is {slightly} improved.
Figure \ref{fig:cifar100_training} shows the classification accuracy of training and test samples in CIFAR100 during training.
As shown {in} the figure,
$\mathcal{L}_\text{DA}$ causes a higher generalization error than others
because $\mathcal{L}_\text{DA}$ forces the unnecessary invariant property.
Moreover, optimizing $\mathcal{L}_\text{MT}$ is harder than doing $\mathcal{L}_\text{SLA}$
as described in Section \ref{sec:method:joint}, thus the former achieves the lower accuracy on both training and test
samples than the latter.
These results show that learning invariance to some transformations, e.g., rotation, makes
optimization harder and degrades the performance. Namely,
such transformations should be carefully handled.

\begin{table*}[t]
\caption{Average classification accuracy (\%) with 95\% confidence intervals of 1000 5-way few-shot tasks
on mini-ImageNet, CIFAR-FS, and FC100.
$\dagger$ and  $\ddagger$ indicates 4-layer convolutional and 28-layer residual networks \cite{zagoruyko2016wrn}, respectively. Others use 12-layer residual networks as 
\citet{lee2019metaoptnet}.
We follow the same experimental settings as
\citet{lee2019metaoptnet} did.
The best accuracy is indicated as bold.
}\label{table:few}
\vskip 0.1in
\begin{center}
\small
\begin{tabular}{ccccccc}
\toprule
& \multicolumn{2}{c}{mini-ImageNet} & \multicolumn{2}{c}{CIFAR-FS} & \multicolumn{2}{c}{FC100} \\
\cmidrule(lr){2-3}\cmidrule(lr){4-5}\cmidrule(lr){6-7}
Method                   & 1-shot           & 5-shot           & 1-shot           & 5-shot           & 1-shot           & 5-shot           \\
\midrule
MAML$^\dagger$ \cite{finn2017maml} & \ms{48.70}{1.84} & \ms{63.11}{0.92} & \ms{58.9}{1.9} & \ms{71.5}{1.0} & -              & - \\
R2D2$^\dagger$ \cite{bertinetto2018r2d2} & -                & -                & \ms{65.3}{0.2} & \ms{79.4}{0.1} & -              & - \\
RelationNet$^\dagger$ \cite{sung2018relationnet} & \ms{50.44}{0.82} & \ms{65.32}{0.70} & \ms{55.0}{1.0} & \ms{69.3}{0.8} & -              & - \\
SNAIL \cite{mishra2018snail}                    & \ms{55.71}{0.99} & \ms{68.88}{0.92} & - & - & - & - \\
TADAM \cite{oreshkin2018tadam}               & \ms{58.50}{0.30} & \ms{76.70}{0.30} & -            & -                & \ms{40.1}{0.4} & \ms{56.1}{0.4} \\
LEO$^\ddagger$ \cite{rusu2018leo} & \ms{61.76}{0.08} & \ms{77.59}{0.12} & - & - & - & - \\
MetaOptNet-SVM \cite{lee2019metaoptnet}          & \ms{62.64}{0.61} & \ms{78.63}{0.46} & \ms{72.0}{0.7} & \ms{84.2}{0.5} & \ms{41.1}{0.6} & \ms{55.5}{0.6} \\
\midrule
ProtoNet  \cite{snell2017prototypical}               & \ms{59.25}{0.64} & \ms{75.60}{0.48} & \ms{72.2}{0.7} & \ms{83.5}{0.5} & \ms{37.5}{0.6} & \ms{52.5}{0.6} \\
ProtoNet + {\bf SLA+AG} (ours)        & \ms{62.22}{0.69} & \ms{77.78}{0.51} & {\bf \ms{74.6}{0.7}} & {\bf \ms{86.8}{0.5}} & \ms{40.0}{0.6} &  \ms{55.7}{0.6} \\
\midrule
MetaOptNet-RR \cite{lee2019metaoptnet}           & \ms{61.41}{0.61} & \ms{77.88}{0.46} & \ms{72.6}{0.7} & \ms{84.3}{0.5} & \ms{40.5}{0.6} & \ms{55.3}{0.6} \\
MetaOptNet-RR + {\bf SLA+AG} (ours)   & {\bf \ms{62.93}{0.63}} & {\bf \ms{79.63}{0.47}} &  \ms{73.5}{0.7} &  \ms{86.7}{0.5} & {\bf \ms{42.2}{0.6}} & {\bf \ms{59.2}{0.5}} \\
\bottomrule
\end{tabular}%}
\end{center}
\vspace{-0.2in}
\end{table*}

\textbf{Comparison with independent ensemble.} Next, to evaluate the effect of the aggregation in SLA-trained models,
we compare the aggregation using rotation with independent ensemble (IE)
which aggregates the {pre-softmax activations (i.e., logits)}
over independently trained models.\footnote{In the supplementary material, we also compare our method with ten-crop \cite{krizhevsky2012alexnet}.}
We here use four independent models (i.e., $4\times$ more parameters than ours) since IE with four models and SLA+AG have the same inference cost.
Surprisingly, as reported in Table \ref{table:compare}, the aggregation using rotation
achieves competitive performance
compared to the ensemble. %with 4 independently trained models.
When using both IE and SLA+AG with rotation, i.e., the same number of parameters as the ensemble,
the accuracy is improved further.

\subsection{Evaluation on Standard Setting}\label{sec:exp:standard}

We demonstrate the effectiveness of our self-supervised augmentation method on various image classification datasets:
CIFAR10/100, CUB200, MIT67, Stanford Dogs, and tiny-ImageNet.
We first evaluate the effect of aggregated inference $P_\text{aggregated}(\cdot|\vx)$
in (\ref{eq:agg}) of Section \ref{sec:method:joint}: see the SLA+AG column in Table \ref{table:all}.
Using rotation as augmentation improves the classification accuracy on all datasets, e.g., 8.60\% and 18.8\% relative
gain over baselines on CIFAR100 and CUB200, respectively.
With color permutation, the performance improvements are less significant on CIFAR and tiny-ImageNet,
but it still provides meaningful gains on fine-grained datasets, e.g., 12.6\% and 10.6\% relative gain on CUB200 and Stanford Dogs, respectively. In the supplementary material, we also provide additional experiments on large-scale datasets, e.g.,
iNaturalist \cite{van2018inaturalist} of 8k labels, to demonstrate the scalability of SLA with respect to the number of labels.

{Since both transformations are effective on the fine-grained datasets,
we also test composed transformations of the two different types of transformations for further improvements.
To construct the composed ones, we first choose two subsets $T_r$ and $T_c$ of rotation and color permutation, respectively,
e.g., $T_r=\{0^\circ,180^\circ\}$ or $T_c=\{\text{RGB}, \text{GBR}, \text{BRG}\}$. Then, we
compose them, i.e., $T=\{t_c\circ t_r:t_r\in T_r,t_c\in T_c\}$.
It means that $t=t_c\circ t_r\in T$ rotates an image by $t_r$
and then swaps color channels by $t_c$.
As reported in Table \ref{table:comb_train}, using a larger set $T$ improves the aggregation inference
further.
However, under too many transformations,
the aggregation performance can be degraded since the optimization becomes too harder.
When using $M=12$ transformations, we achieve the best performance, $20.8\%$ relatively higher than the baseline on CUB200. Similar experiments on Stanford Dogs are reported in the supplementary material.}

We further apply SLA+SD (that is faster than SLA+AG in inference) with existing augmentation techniques, Cutout \cite{devries2017cutout},
CutMix \cite{yun2019cutmix}, AutoAugment \cite{cubuk2018autoaugment}, and FastAutoAugment \cite{lim2019fastautoaugment} into recent
architectures \cite{zagoruyko2016wrn,han2017pyramidnet}.
Note that SLA uses semantically-sensitive transformations for assigning different labels, while
conventional data augmentation methods use semantically-invariant transformations for preserving
labels. Thus, transformations using SLA and conventional data augmentation (DA) techniques do not overlap. For example,
the AutoAugment \cite{cubuk2018autoaugment} policy rotates images at most $30$ degrees, while SLA does at least $90$ degrees. Therefore, SLA can be naturally combined with the existing DA methods.
As reported in Table \ref{table:cifar-sota}, SLA+SD consistently reduces the classification
errors. As a result, it achieves $1.80\%$ and $12.24\%$ error rates on CIFAR10/100, respectively.
These results demonstrate the compatibility of the proposed method.

\begin{table*}[t]
\caption{Classification accuracy (\%) on imbalanced datasets of CIFAR10/100.
Imbalance Ratio is the ratio between the numbers of samples of most and least frequent classes.
We follow the experimental settings of
\citet{cao2019ldam}.
The best accuracy is indicated as bold, and we use brackets to report the relative accuracy gains over {each counterpart that does not use SLA}.
}\label{table:im}
\vskip 0.1in
\begin{center}
\small
%\resizebox{\textwidth}{!}{%
\begin{tabular}{ccccc}
\toprule
                               & \multicolumn{2}{c}{Imbalanced CIFAR10}                         & \multicolumn{2}{c}{Imbalanced CIFAR100}                        \\
                                 \cmidrule(lr){2-3}                                                \cmidrule(lr){4-5}            
Imbalance Ratio ($N_\text{max}/N_\text{min}$)                
                               & 100                      & 10                                   & 100                            & 10                             \\
\midrule
Baseline                            &      70.36                     &      86.39                     &      38.32                     &      55.70                     \\
Baseline + {\bf SLA+SD} (ours)      &  74.61 {\small (+6.04\%)} &  89.55 {\small (+3.66\%)} &  43.42 {\small (+13.3\%)} &  60.79 {\small (+9.14\%)} \\
\midrule                                                                                                                                                           
CB-RW \cite{cui2019cb}        &      72.37                     &      86.54                     &      33.99                     &      57.12                     \\
CB-RW + {\bf SLA+SD} (ours)    & 77.02 {\small (+6.43\%)} &  89.50 {\small (+3.42\%)} &  37.50 {\small (+10.3\%)} & {\bf 61.00} {\small (+6.79\%)} \\
\midrule                                                                                                                                                           
LDAM-DRW \cite{cao2019ldam}   &      77.03                     &      88.16                     &      42.04                     &      58.71                     \\
LDAM-DRW + {\bf SLA+SD} (ours) & {\bf 80.24} {\small (+4.17\%)} & {\bf 89.58} {\small (+1.61\%)} & {\bf 45.53} {\small (+8.30\%)} & 59.89 {\small (+1.67\%)} \\
    \bottomrule
\end{tabular}%}
\end{center}
\vspace{-0.05in}
\end{table*}

\subsection{Evaluation on Limited-data Setting}

\begin{figure}
\centering
\includegraphics[width=0.35\textwidth]{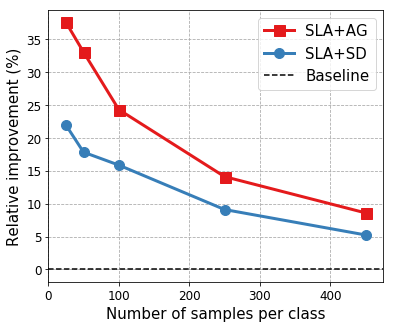}
\caption{Relative improvements (\%) over baselines
under varying {the number of training samples per class in CIFAR100.}
}\label{fig:cifar100_limit}
\vspace{-0.05in}
\end{figure}

\textbf{Limited-data regime.} Our augmentation techniques are also effective when only few training samples are available.
To evaluate the effectiveness,
we first construct sub-datasets of CIFAR100 via randomly choosing $n\in\{25,$ $50,$ $100,$ $250\}$ samples for each class,
and then train models with and without our rotation-based self-supervised label augmentation. As shown in Figure \ref{fig:cifar100_limit},
our scheme improves the accuracy relatively up to 37.5\% under aggregation
and 21.9\% without aggregation.

\textbf{Few-shot classification.} 
Motivated by the above results in the limited-data regime,
we
also apply our SLA+AG\footnote{In few-shot learning, it is hard to define the additional classifier
$\sigma(f(\vx;\vtheta);\vu)$ in \eqref{eq:loss_sda_sd}
for
unseen classes when applying SLA+SD.}
method to solve few-shot classification,
combined with
recent approaches, ProtoNet \cite{snell2017prototypical} and
MetaOptNet \cite{lee2019metaoptnet} specialized for this problem.
Note that our method augments $N$-way $K$-shot tasks to $NM$-way $K$-shot when
using $M$-way transformations. As reported in Table \ref{table:few}, ours improves consistently 5-way 1/5-shot classification
accuracy on mini-ImageNet, CIFAR-FS, and FC100. For example, we obtain 7.05\% relative improvements on 5-shot tasks of FC100.
Here, we remark that one may obtain further improvements by applying
additional data augmentation techniques to ours (and the baselines), as we shown in Section \ref{sec:exp:standard}.
However, we found that
training with the state-of-the-art data augmentation technique
and/or testing with ten-crop \cite{krizhevsky2012alexnet} do not always provide meaningful improvements for the few-show experiments, e.g., the AutoAugment~\cite{cubuk2018autoaugment} policy
and ten-crop provide marginal ($<$1\%) accuracy gain on FC100 under ProtoNet in our experiments.

\textbf{Imbalanced classification.} 
Finally, we consider a setting of imbalanced training datasets, where the number of instances per class largely differs and some classes have only a few training instances.
For this experiment, we combine our SLA+SD method
with two recent approaches, the Class-Balanced (CB) loss \cite{cui2019cb}
and LDAM \cite{cao2019ldam}, specialized for this problem.
Under imbalanced datasets of CIFAR10/100, which have long-tailed label distributions,
our approach consistently improves the classification accuracy %on the imbalance datasets
as reported in Table \ref{table:im} (e.g., up to $13.3\%$ relative
gain on an imbalanced CIFAR100 dataset).
The results show the wide applicability of our self-supervised label augmentation.
Here, we emphasize that all tested methods (including our SLA+SD) have the same inference time.

% \vspace{-0.05in}
\section{Related Work}

\textbf{Self-supervised learning.}
For representation learning in unlabeled datasets,
self-supervised learning approaches construct
artificial labels (referred to as self-supervision)
using only input signals, and then learn to predict them.
The self-supervision can be constructed in various ways.
A simple one of them is transformation-based approaches~\cite{doersch2015context,noroozi2016jigsaw,larsson2017colorization,gidaris2018rotation,zhang2019aet}.
They first modify inputs by a transformation, e.g., rotation \cite{gidaris2018rotation} and
patch permutation \cite{noroozi2016jigsaw}, and then assign
the transformation as the input's label. 

Another approach is clustering-based ~\cite{bojanowski2017predicting_noise,caron2018deepcluster,wu2018instance_discrimination,ym.2020selflabelling}. They first perform
clustering using the current model, and then assign labels using
the cluster indices. When performing this procedure iteratively, 
the quality of representations is gradually improved.
Instead of clustering,
\citet{wu2018instance_discrimination} assign different labels for each sample, i.e., consider each sample as a cluster.

% the simpl
% The construction strategies for self-supervision
% fall into three categories: clustering-based~\cite{caron2018deepcluster,ym.2020selflabelling} and instance discrimination~\cite{}.
% The transformation-based approaches construct labels by input transformations, e.g., rotation \cite{gidaris2018rotation} and
% patch permutation \cite{noroozi2016jigsaw}. They first modify inputs
% by a transformation, and assign its label as the transformation.
% On the other hand,
% the clustering-based approaches construct labels by clustering
% first modifies inputs by a transformation such as rotation, and construct 

While the recent clustering-based approaches
outperform transformation-based ones for unsupervised learning,
the latter is widely used for other purposes due to its simplicity, e.g., semi-supervised
learning \cite{zhai2019s4l,berthelot2020remixmatch}, 
improving robustness \cite{hendrycks2019self_robustness},
and training generative adversarial networks
\cite{chen2018self_gan}.
In this paper, we also utilize transformation-based self-supervision, but aim to improve accuracy under full-supervised
datasets.

% For learning deep n
% In realistic scenarios,

\textbf{Self-distillation.}
\citet{hinton2015knowledge_distillation} 
propose a 
knowledge distillation
technique,
which improves a network via transferring (or distilling)
knowledge of a pre-trained larger network.
There are many advanced distillation
techniques \cite{zagoruyko2016at,park2019rkd,ahn2019vid,Tian2020crd},
but they should train the larger network first,
which leads to high training costs.
To overcome this shortcoming,
self-distillation approaches, which transfer
own knowledge into itself, have been developed.
\cite{lan2018one,zhang2019your,xu2019data}.
They utilize partially-independent architectures \cite{lan2018one},
data distortion \cite{xu2019data},
or hidden layers \cite{zhang2019your} for distillation.
While these approaches perform distillation on the same label space, our framework
transfers knowledge between different label spaces
augmented by self-supervised transformations.
Thus our approach could enjoy an orthogonal usage with the existing ones;
for example, one can distill aggregated knowledge $P_\text{aggregated}$ \eqref{eq:agg}
into hidden layers as \citet{zhang2019your} did.

% thus we do 
% does not require such techniques because
% Moreover, our distillation method could be extended by 
% Compared to these approaches, ours 
% For example, \citet{lan2018one} distill ensemble knowledge of
% a partially-independent architecture in an online manner,
% \citet{zhang2019your} transfer  into hidden layers,
% and \citet{xu2019data} transfer between differently distorted samples.
% Compared to these approaches, our framework uses only
% one model, and different labels
% So our method can be extended as layer-wise  that 

% However, they 

% partially-separated models in an
% online manner
% use 
% for producing an ensemble prediction while ours does not.
% As reported in \table{xx}, when 
% \cite{xu2019data} 
% They utilize
% partially-independent architectures for ensemble \cite{lan2018one}
% or
% data distortion \cite{xu2019data} for distillation.
% producing an ensemble prediction of diverse outputs

% distill knowledge of a current model into the model itself.

% \citet{lan2018one} distill
% ensemble knowledge of partially-separated models in an
% online manner. However, they still need separated models/branches
% because ensemble requires diverse predictions.
% Compared to this approach, in our framework, a single model
% can provide diverse predictions for ensemble because we augment
% labels by transformations.

\section{Conclusion}\label{sec:conclusion}
We proposed a simple yet effective approach utilizing self-supervision on fully-labeled datasets via
learning a single unified task with respect to the joint distribution of the original and self-supervised labels.
We think that our work could bring in many interesting directions for future research; for instance, one can revisit prior works on applications of self-supervision, e.g., 
semi-supervised learning with self-supervision~\cite{zhai2019s4l,berthelot2020remixmatch}. Applying our joint learning framework to fully-supervised tasks other than the few-shot or imbalanced classification task, or learning to select tasks that are helpful toward improving the main task prediction accuracy, are other interesting future research directions. 

\section*{Acknowledgements}

This work was partly supported by Institute of Information \& Communications Technology Planning \& Evaluation (IITP) grant funded by the Korea government (MSIT) (No.2019-0-00075, Artificial Intelligence Graduate School Program (KAIST)). This work was mainly supported by
Samsung Research Funding \& Incubation Center of Samsung Electronics under Project Number SRFC-IT1902-06.

\bibliography{main}
\bibliographystyle{icml2020}

\clearpage
\onecolumn
%\begin{center}{\bf {\LARGE Supplementary Material}}
%\end{center}
%\begin{center}{\bf {\Large Self-supervised Label Augmentation via Input Transformations}}
%\end{center}

\appendix

\section{Comparison with ten-crop}

In this section, we compare the proposed aggregation method (SLA+AG) using rotation with a widely-used aggregation scheme, ten-crop \cite{krizhevsky2012alexnet},
which aggregates the pre-softmax activations (i.e., logits) over a number of cropped images.
As reported in Table \ref{table:ten_crop},
the aggregation using rotation performs significantly better than ten-crop.

\begin{table*}[h]
\vspace{-0.1in}
\caption{Classification accuracy (\%) of the ten-crop and our aggregation using rotation (SLA+AG).
The best accuracy is indicated as bold, and the relative gain over the baseline is shown in brackets.}\label{table:ten_crop}
\begin{center}
\small
\begin{tabular}{ccccc}
\toprule
Dataset       & Baseline & ten-crop & SLA+AG \\
\midrule                                                    
CIFAR10       & 92.39    & 93.33  {\small (+1.02\%)}  & {\bf 94.50} {\small (+2.28\%)} & \\
CIFAR100      & 68.27    & 70.54  {\small (+3.33\%)}  & {\bf 74.14} {\small (+8.60\%)} & \\
tiny-ImageNet & 63.11    & 64.95  {\small (+2.92\%)}  & {\bf 66.95} {\small (+6.08\%)} & \\ % \midrule
\bottomrule
\end{tabular}
\end{center}
\vspace{-0.1in}
\end{table*}

\section{Experiments with Composed Transformations}

In this section, we present the more detailed experimental results of composed transformations
described in the main text (Section 3.3 and Table 4). We additionally report the performance
of the single inference (SLA+SI) with an additional dataset, Stanford Dogs.
When using $M=12$ composed transformations, we achieve the best performance, $20.8\%$ and $15.4\%$ relatively higher than baselines on CUB200 and Stanford Dogs, respectively.

\begin{table*}[h]
\vspace{-0.1in}
\caption{
Classification accuracy (\%) of SLA based on the set (each row) of composed transformations.
We first choose subsets of rotation and color permutation (see first two columns) and compose them
where $M$ is the number of composed transformations.
We use SLA with the composed transformations when training models.
The best accuracy is indicated as bold.}\label{table:comb_train}
\begin{center}
\small
%\resizebox{\textwidth}{!}{
\begin{tabular}{ccrcccc}
\toprule
\multicolumn{2}{c}{{Composed transformations $T=T_r\times T_c$}} & & \multicolumn{2}{c}{CUB200} & \multicolumn{2}{c}{Stanford Dogs} \\
\cmidrule(lr){1-2}\cmidrule(lr){4-5}\cmidrule(lr){6-7}
Rotation {$T_r$}                                & Color permutation {$T_c$}    & $M$ & SLA+SI & SLA+AG & SLA+SI & SLA+AG \\
\midrule
0$^\circ$                                       & RGB                          & 1   & \multicolumn{2}{c}{54.24} & \multicolumn{2}{c}{60.62} \\
0$^\circ$, 180$^\circ$                          & RGB                          & 2   &      56.62  &      58.92  &      63.57  &      65.65  \\
0$^\circ$, 90$^\circ$, 180$^\circ$, 270$^\circ$ & RGB                          & 4   &      60.85  &      64.41  &      65.67  &      67.03  \\
0$^\circ$                                       & RGB, GBR, BRG                & 3   &      52.91  &      56.47  &      63.26  &      65.87  \\
0$^\circ$                                       & RGB, RBG, GRB, GBR, BRG, BGR & 6   &      56.81  &      61.10  &      64.83  &      67.03  \\
0$^\circ$, 180$^\circ$                          & RGB, GBR, BRG                & 6   &      56.14  &      60.87  &      65.45  &      68.75  \\
0$^\circ$, 90$^\circ$, 180$^\circ$, 270$^\circ$ & RGB, GBR, BRG                & 12  &      60.74  & {\bf 65.53} & {\bf 66.40} & {\bf 69.95} \\
0$^\circ$, 90$^\circ$, 180$^\circ$, 270$^\circ$ & RGB, RBG, GRB, GBR, BRG, BGR & 24  & {\bf 61.67} &      65.43  &      64.71  &      67.80  \\
\bottomrule
\end{tabular}%}
\end{center}
\vspace{-0.1in}
\end{table*}

\newpage
\section{Self-supervised Label Augmentation with Thousands of Labels}

Since the proposed technique (SLA) increases
the number of classes in a task, one could
wonder that the technique is scalable with respect to the number of labels.
To demonstrate the scalability of SLA, we train ResNet-50 \cite{he2016resnet} on
ImageNet \cite{deng2009imagenet} and iNaturalist \cite{van2018inaturalist} datasets
%ImageNet-128$\times$128 \cite{chrabaszcz2017downsampled},
%a downsampled version of ImageNet \cite{deng2009imagenet},
with the same
experimental settings of tiny-ImageNet as described in the main text (Section 3.1) except the number of training iterations. In this experiment, we train models for 900K and 300K iterations
(roughly 90 epochs) for ImageNet and iNaturalist, respectively. 
%Note that the dataset has 1K classes and 1.2M training images while tiny-ImageNet has 200 classes and 100K training images.
As reported in Table \ref{table:imagenet128},
our method also provides a benefit on the large-scale datasets.

\begin{table*}[h]
\vspace{-0.1in}
\caption{Classification accuracy (\%) on ImageNet \cite{deng2009imagenet} and iNaturalist \cite{van2018inaturalist} with SLA using rotation. $N$ indicates the number of labels in each dataset.
The relative gain over the baseline is shown in brackets. Note that the reported accuracies are obtained from only one trial.}\label{table:imagenet128}
\begin{center}
\small
\begin{tabular}{cccccc}
\toprule
Dataset       & $N$ & Baseline & SLA+SI & SLA+AG & SLA+SD \\
\midrule
%ImageNet-128$\times$128 & 61.34    & 63.20  {\small (+3.03\%)}  & 66.20 {\small (+7.92\%)} & 64.21 {\small (+4.68\%)} \\
ImageNet \cite{deng2009imagenet} & 1000
& 75.16 & 75.81 {\small (+0.86\%)} & 77.16 {\small (+2.66\%)} & 76.17 {\small (+1.34\%)} \\
iNaturalist \cite{van2018inaturalist} & 8142
& 57.12 & 61.31 {\small (+7.34\%)} & 62.97 {\small (+10.2\%)} & 61.52 {\small (+7.70\%)} \\
\bottomrule
\end{tabular}
\end{center}
\vspace{-0.1in}
\end{table*}

\section{Combining with A Self-supervised Pre-training Technique}

While self-supervised learning (SSL) techniques primarily target unsupervised learning or pre-training, we focus on joint-supervised-learning (from scratch) with self-supervision to improve upon the original supervised learning model. Thus our SLA framework is essentially a supervised learning method, and is not comparable with SSL methods that train with unlabeled data.

Yet, since our SLA is orthogonal from SSL, we could use a SSL technique as a pre-training strategy for our scheme as well. To validate this, we pre-train ResNet-18 \cite{he2016resnet} on CIFAR-10 \cite{krizhevsky2009cifar} using a SOTA contrastive learning method, SimCLR \cite{chen2020simclr}, and then fine-tune it on the same dataset. As reported in Table \ref{table:pretraining}, under the fully-supervised settings, pre-training the network only with SimCLR yields a marginal performance gain. On the other hand, when using SimCLR for pre-training and ours for fine-tuning, we achieve a significant performance gain, which shows that the benefit of our approach is orthogonal to pre-training strategies.

\begin{table*}[h]
\vspace{-0.1in}
\caption{Classification accuracy (\%) on CIFAR-10 \cite{krizhevsky2009cifar} with SimCLR \cite{chen2020simclr} and our SLA framework.}\label{table:pretraining}
\begin{center}
\small
\begin{tabular}{ccc}
\toprule
Initialization & Fine-tuning & Accuracy (\%) \\ \midrule
\multirow{2}{*}{Random initialization} & Baseline      & 95.26 \\
                                       & SLA+SD (ours) & 96.19 \\ \midrule
\multirow{2}{*}{SimCLR pre-training}   & Baseline      & 95.44 \\
                                       & SLA+SD (ours) & 96.55 \\
\bottomrule
\end{tabular}
\end{center}
\vspace{-0.1in}
\end{table*}

\newpage
\section{Implementation with PyTorch}

One of the strengths of the proposed self-supervised label augmentation is simple to implement. Note that
the joint label $y=(i,j)\in[N]\times[M]$ can be rewritten as a single label $y=M\times i + j$ where $N$ and $M$ are
the number of primary and self-supervised labels, respectively. 
Thus self-supervised label augmentation (SLA) can be implemented in PyTorch as follows.
Note that \verb+torch.rot90(X, k, ...)+ is a built-in function which rotates the input tensor $X$ by $90k$ degrees.

\lstset{
language=Python,
numbers=left,
breaklines=true,
xleftmargin=5.0ex,
basicstyle=\footnotesize\ttfamily
}
    
\begin{lstlisting}[caption=Training script of self-supervised label augmentation.,label={listing:training}]
for inputs, targets in train_dataloader:
    inputs = torch.stack([torch.rot90(inputs, k, (2, 3)) for k in range(4)], 1)
    inputs = inputs.view(-1, 3, 32, 32)
    targets = torch.stack([targets*4+k for k in range(4)], 1).view(-1)
    
    outputs = model(inputs)
    loss = F.cross_entropy(outputs, targets)
    
    optimizer.zero_grad()
    loss.backward()
    optimizer.step()
\end{lstlisting}

\begin{lstlisting}[caption=Evaluation script of self-supervised label augmentation with single (SLA+SI) and aggregated (SLA+AG) inference.,label={listing:eval}]
for inputs, targets in test_dataloader:
    outputs = model(inputs)
    SI = outputs[:, ::4]

    inputs = torch.stack([torch.rot90(inputs, k, (2, 3)) for k in range(4)], 1)
    inputs = inputs.view(-1, 3, 32, 32)
    outputs = model(inputs)
    AG = 0.
    for k in range(4):
        AG = AG + outputs[k::4, k::4] / 4.

    SI_accuracy = compute_accuracy(SI, targets)
    AG_accuracy = compute_accuracy(AG, targets)
\end{lstlisting}

As described above, applying input transformations (e.g., line 2-3 in Listing \ref{listing:training}) and label augmentation (e.g., line 4 in Listing \ref{listing:training}) is enough to implement SLA.
We here omit the script for SLA with self-distillation (SLA+SD), but remark that its implementation is also simple as SLA.
We think that this simplicity could lead to the broad applicability for various applications.

%\bibliography{main}
%\bibliographystyle{icml2020}

%\end{document}

% This document was modified from the file originally made available by
% Pat Langley and Andrea Danyluk for ICML-2K. This version was created
% by Iain Murray in 2018, and modified by Alexandre Bouchard in
% 2019 and 2020. Previous contributors include Dan Roy, Lise Getoor and Tobias
% Scheffer, which was slightly modified from the 2010 version by
% Thorsten Joachims & Johannes Fuernkranz, slightly modified from the
% 2009 version by Kiri Wagstaff and Sam Roweis's 2008 version, which is
% slightly modified from Prasad Tadepalli's 2007 version which is a
% lightly changed version of the previous year's version by Andrew
% Moore, which was in turn edited from those of Kristian Kersting and
% Codrina Lauth. Alex Smola contributed to the algorithmic style files.

\end{document}